# WAT: Wave-Attractor-Tree - A Hierarchical Binary Tree Architecture for Efficient Sequence Modeling

**Igor Berezkin**

*Independent Researcher*

2025

## Abstract

We present WAT (Wave-Attractor-Tree), a neural architecture for sequence modeling that replaces self-attention with a hierarchical binary tree reduction. WAT requires O(n) total merge operations with O(log n) parallel depth and O(n) space per level, in contrast to the O(n²) complexity of standard Transformer self-attention. The core mechanism is a Gated Linear Unit (GLU) merge operation with RMSNorm normalization, applied recursively at each level of a binary tree over the input token sequence. We evaluate three progressively refined variants of WAT against a Transformer baseline with matched parameter counts (~106K parameters each) on character-level language modeling (TinyShakespeare, sequence length 512) and bracket balance classification. WAT V1 (One-to-One, tree reduction to root) achieves 45.10% next-character accuracy versus 42.83% for the Transformer, training 10× faster per epoch (10s vs. 100s). WAT V2 (Seq2Seq with causal prefix scan) achieves 47.29% accuracy versus 36.28% for the Transformer (+11.0 pp), demonstrating that dense supervision amplifies the hierarchical advantage. WAT V3 (Seq2Seq with chunk-based parallel tree reduction) matches V2 accuracy at 47.21% while recovering V1 training speed (~9s/epoch), resolving the speed-accuracy trade-off. On bracket balance classification with long sequences (512-1024 tokens), WAT (tree reduction, no wave) achieves 75.0% accuracy versus 57.0% for the Transformer - an 18.0 percentage point gap - with WAT training 10× faster per epoch (~1.1s vs. ~11s). WAT-Chunk (chunk-based) achieves 55.0% on this task, matching the Transformer, while WAT+Wave achieves 74.8% - confirming that full tree reduction (not chunk-based approximation) is critical for structural classification tasks. Code, model weights, and all experimental logs are released publicly.

**Open Source Implementation:** The complete source code for WAT V1, V2, and V3, along with training logs and pre-trained configurations, is available at: https://github.com/IgorBerezkin/WAT

## 1. INTRODUCTION

The Transformer architecture [1] has become the dominant paradigm for sequence modeling, underpinning state-of-the-art systems in natural language processing, protein structure prediction, and beyond. However, the self-attention mechanism at the Transformer's core computes pairwise interactions between all tokens, resulting in O(n²) time and memory complexity. This quadratic scaling becomes a fundamental bottleneck as sequence lengths grow: doubling the sequence length quadruples the compute and memory required for attention. At sequence length 512, a single attention layer computes 512² = 262,144 scalar products; at length 4096, this grows to over 16 million.

This paper proposes WAT (Wave-Attractor-Tree), a conceptually simple alternative that replaces global self-attention with a hierarchical binary tree reduction. The core insight is that rich sequence representations can be built bottom-up rather than all-at-once: adjacent pairs of token embeddings are merged via a learned Gated Linear Unit (GLU) operation, producing a set of higher-level representations half the size of the original. This merging is applied recursively, producing a logarithmically deep hierarchy. Each level of the hierarchy is fully parallelizable - all merges at level l are independent - giving O(log n) sequential depth and O(n log n) total work.

We develop three variants of WAT across this paper. WAT V1 (One-to-One) reduces the entire past sequence to a single root vector and predicts the next token. WAT V2 (Seq2Seq with Causal Scan) produces a contextual representation for every position through a causal prefix scan. WAT V3 (Seq2Seq with Chunk-Based Tree Reduction) achieves the accuracy of V2 with the training speed of V1 through a key architectural insight: by partitioning the sequence into fixed-size chunks and computing tree summaries in parallel across all chunks, we eliminate sequential dependencies while preserving causality.

Our main contributions are:
- A hierarchical binary tree architecture (WAT) with O(n log n) computational complexity for autoregressive sequence modeling.
- A GLU + RMSNorm merge operation with residual gating, applied recursively with shared weights across all tree levels.
- A chunk-based Seq2Seq formulation (WAT V3) achieving O(n log K) complexity where K is the chunk size, with full GPU parallelism and strict causal guarantees.
- Empirical demonstration that WAT outperforms a matched Transformer baseline by +11 pp on language modeling and +18 pp on long-sequence bracket classification, training 10× faster per epoch on the classification task.

## 2. RELATED WORK

### 2.1 Efficient Transformers

The quadratic cost of self-attention has motivated extensive research on efficient alternatives. Sparse Transformer [2] applies attention to sparse subsets of positions (local windows and strided global patterns), reducing complexity to $O(n\sqrt{n})$. Longformer [3] combines windowed local attention with task-specific global tokens. Linformer [4] projects keys and values to low rank, achieving O(n) attention at the cost of expressiveness. Performer [5] approximates softmax attention via random Fourier features. BigBird [6] combines local, global, and random attention patterns with theoretical guarantees.

These approaches share a common strategy: modifying or approximating the attention computation while preserving its fundamental structure. WAT takes a different path, replacing attention entirely with a tree-structured aggregation that has no attention matrix at all.

## 2.2 State-Space and Recurrent Models

S4 [7] models long-range dependencies using structured state-space models (SSMs), achieving O(n log n) computation via convolution in the frequency domain. Mamba [8] extends SSMs with input-selective gating, achieving competitive language modeling performance at linear inference cost. RWKV [9] reformulates Transformer attention as a linear recurrence, enabling RNN-style O(1) inference while training like a Transformer. Hyena [10] uses long convolutions as implicit attention substitutes.

WAT differs from SSM-based models in that it is non-recurrent and non-convolutional. Unlike S4 or Mamba, WAT does not process positions sequentially - it applies the same merge operation in parallel at each tree level. This makes WAT naturally suited for GPU parallelism and eliminates the hidden-state bottleneck of recurrent models.

## 2.3 Tree-Structured Models

TreeLSTM [11] applies LSTM units over explicit syntactic parse trees, achieving strong results on semantic tasks. Recursive neural networks [12] apply composition functions over constituency parse trees. These models require external structural supervision (parse trees), whereas WAT operates on a fixed balanced binary tree topology determined only by sequence length, requiring no structural annotations.

Closest in spirit to WAT is the parallel prefix scan [13], a classical algorithm for parallel computation of cumulative operations. WAT's tree reduction is related to the reduce phase of prefix scan, and the Seq2Seq causal variant (V2) is structurally analogous to the scheduling pattern of a causal prefix scan with a learned merge operation. The analogy concerns the communication pattern and doubling schedule rather than strict algebraic associativity.

Beyond syntactic TreeLSTM models, several works have explored hierarchical or latent tree structures learned directly from data, including latent tree induction models and parallel tree encoding approaches. Unlike such models, WAT does not learn tree topology and does not rely on external syntactic supervision. Instead, it operates on a fixed balanced binary tree determined solely by sequence length, with weight sharing across levels.

## 2.4 Gated Linear Units and Normalization

Gated Linear Units (GLU) [14] were introduced as multiplicative gating in convolutional language models. $GLU(x) = (W_{1X} + b_1) \odot \sigma(W_{2X} + b_2)$ provides selective information flow and has proven effective in Transformer feed-forward layers (SwiGLU [15], GeGLU). WAT applies GLU-style gating at every level of the binary tree merge, controlling information flow between sibling subtrees. RMSNorm [16], applied after each GLU merge, provides training stability across the O(log n) sequential merge levels without the computational overhead of full LayerNorm.

# 3. METHODOLOGY

## 3.1 Problem Setup

We consider autoregressive language modeling: given a sequence of discrete tokens $x = (x_1, x_2, ..., x_n)$, model the conditional distribution $p(x_t | x_1, ..., x_{t-1})$ for each position t. At inference time, this allows generation of new sequences one token at a time.

## 3.2 Input Encoding (Common to All Variants)

All WAT variants share a common input encoding pipeline:

Token Embedding: each token $x_t$ is mapped to a d-dimensional vector via a learned embedding matrix $E \in \mathbb{R}^{|V| \times d}$. Positional encodings $P \in \mathbb{R}^{n \times d}$ are added element-wise:

$$e_t = E[x_t] + P[t] \qquad e_t \in \mathbb{R}^d$$

Causal Convolution: a Conv1d layer with kernel size 3 and left-only padding (padding = kernel_size - 1 on the left, 0 on the right) captures local n-gram context without leaking future tokens. For V1 this is a standard Conv1d with padding=1; for V2/V3 this is an explicit CausalConv1d module:

$$c_t = Conv1d_{casual(e)_t} \qquad (\text{sees only } e_{\{t-2\}}, e_{\{t-1\}}, e_t)$$

Input Gate: an element-wise multiplicative gate filters the convolution output adaptively:

$$n_t = c_t \odot \sigma(W_{gate} \cdot c_t) \qquad W_{gate} \in \mathbb{R}^{d \times d}$$

The sequence $N = (n_1, ..., n_n)$ of gated node representations forms the input to the tree reduction.

## 3.3 Core Operation: Binary Tree Reduction

The fundamental building block of WAT is the binary tree reduction. Given a sequence of d-dimensional vectors $(h_1, ..., h_n)$, adjacent pairs are merged level by level until a single root vector remains.

### 3.3.1 Pairwise Merge Operation

At each level, even- and odd-indexed nodes are separated and concatenated:

$$left = (h_1, h_3, h_5, ...) \qquad \in \mathbb{R}^{\lfloor \frac{n}{2} \rfloor \times d}$$

$$right = (h_2, h_4, h_6, ...) \qquad \in \mathbb{R}^{\lfloor \frac{n}{2} \rfloor \times d}$$

$$combined = concat(left, right) \qquad \in \mathbb{R}^{\lfloor \frac{n}{2} \rfloor \times 2d}$$

The GLU merge computes:

$$val = W_{val} \cdot combined \qquad W_{val} \in \mathbb{R}^{2d \times d}$$

$$gate = \sigma(W_{gate} \cdot combined) \qquad W_{gate} \in \mathbb{R}^{2d \times d}$$

$$merged = RMSNorm(val \odot gate)$$

A learnable residual gate controls the blend between the learned merge and the arithmetic mean:

$$res_{gate} = \sigma(W_{res} \cdot combined) \qquad W_{res} \in \mathbb{R}^{2d \times d}$$

$$residual = \frac{(left + right)}{2}$$

$$h_{next} = res_{gate} \odot merged + (1 - res_{gate}) \odot residual$$

This residual gate serves two roles: (1) it provides a gradient highway when the learned merge is not yet informative early in training, and (2) it allows the model to interpolate between a rich learned aggregation and a simple averaging, making training more stable.

All three weight matrices (W_val, W_gate, W_res) are shared across all tree levels. This parameter sharing is a form of implicit regularization: the merge operation must generalize to representations at every granularity, from token pairs to multi-thousand-token aggregates. It also keeps parameter count independent of sequence length.

### 3.3.2 Complexity Analysis

The total number of merge operations across all levels is:

$$\sum_{l=1}^{\log_2 n} \frac{n}{2^l} = n - 1 = O(n)$$

The total number of merge operations across all levels is n - 1 = O(n). Each merge applies three linear projections over 2d-dimensional inputs. Therefore, the total computational work is $O(n\,d^2)$. Importantly, the tree has O(log n) sequential levels. Thus, WAT's computation has $O(n\,d^2)$ total work and O(log n) parallel depth. The logarithmic depth determines the number of synchronization steps required on parallel hardware, while the linear work determines overall FLOPs.

## 3.4 WAT V1: One-to-One Next-Token Prediction

For predicting the single next token given a context of n past tokens, WAT V1 applies tree reduction to all past nodes and combines the root with the final token:

```
#                    V1 forward pass (WAT_v1.py)
nodes_past = nodes[:, :-1, :]          # (B, n-1, d) - causal: exclude current
root_seq   = tree_reduction(nodes_past) # (B, 1, d)  - global summary
root       = root_seq[:, 0, :]         # (B, d)
last       = nodes_past[:, -1, :]      # (B, d)     - most recent token
```

```
context    = concat([root, last], -1)    # (B, 2d)
logits     = W_predict(context)          # (B, |V|)
```

The concatenation of root and last serves two complementary roles. The root vector is the hierarchical summary of all past tokens - it captures global structure (topic, speaker, stylistic patterns). The last vector is the most recent token representation - it captures immediate local context (the preceding character or word). This design is deliberately asymmetric: the model receives one global signal and one local signal, forcing it to integrate information across different temporal scales.

V1 limitation: the single root aggregates n-1 tokens into one vector through O(log n) lossy compressions. Information about distant tokens may be partially lost. This is the motivation for the Seq2Seq variants below.

### 3.5 WAT V2: Seq2Seq via Causal Prefix Scan

For sequence-to-sequence prediction (producing a next-token prediction at every position), WAT V2 employs a causal prefix scan: for each position t, the representation h_t should encode only the past tokens $x_1, ..., x_{t-1}$.

It is important to note that the GLU-based merge operation used in WAT is not strictly associative. Therefore, WAT V2's causal scan should be understood as a learned hierarchical aggregation scheme with prefix-like structure, rather than a mathematically exact associative prefix scan. The prefix-scan analogy refers to the communication pattern and doubling schedule, not to algebraic associativity.

The causal prefix scan maintains a running state curr initialized to the input nodes and applies the following update for doubling steps:

```
#                       V2 causal_scan (from WAT_v2.py)
curr = nodes                                 # (B, n, d)
step = 1
while step < n:
    left     = curr[:, :-step, :]            # tokens
    right    = curr[:, step:, :]             # tokens step..n-1
    combined = concat([left, right], dim=-1)
    merged   = GLU_merge(combined)
    new_curr = curr.clone()                  # sequential bottleneck
    new_curr[:, step:, :] = merged
    curr     = new_curr
    step    *= 2                             # steps: 1,2,4,8,...,256
```

After $log_2(n) = 9$ iterations (for n=512), curr[:, t, :] contains the causal prefix summary of tokens 0..t. The output logits are predicted directly from curr:

$$logits_t = W_{predict}(curr_t) \qquad \forall t \in \{1, ..., n\}$$

V2 delivers dense supervision: each forward pass produces n next-token predictions rather than 1, multiplying the training signal by n with no additional data. This accounts for V2's substantially faster convergence and higher peak accuracy compared to V1.

V2 limitation: the clone() operation inside the while loop allocates a fresh O(Bnd) tensor at each of the 9 steps, breaking CUDA parallelism and creating sequential memory allocation overhead. This is the primary cause of V2's ~36s/epoch training time, versus ~10s for V1.

## 3.6 WAT V3: Seq2Seq via Chunk-Based Parallel Tree Reduction

WAT V3 resolves V2's speed bottleneck by replacing the sequential causal scan with a two-stage parallel approach: (1) independent parallel tree reduction over fixed-size chunks, and (2) causal aggregation of chunk summaries via cumulative mean.

### 3.6.1 Chunk Partitioning and Parallel Tree Reduction

The sequence of n tokens is divided into $C = \lceil \frac{n}{K} \rceil$ non-overlapping chunks of size K (default K=32). The key insight is that all chunks can be processed simultaneously by reshaping the batch dimension:

```
#                   V3 chunk_based_forward (from WAT_v3.py)
chunks = nodes.unfold(1, K, K).transpose(2,3)              # (B, C, K, d)

# Parallel tree reduction: reshape treats B*C as independent batch
chunk_summaries = tree_reduction_all(chunks)               # (B, C, d)
```

Inside tree_reduction_all, the (B, C, K, d) tensor is implicitly treated as (B×C, K, d), so all C chunks across all B batch elements are reduced in a single set of operations. There is no clone(), no sequential dependency, and no inter-chunk communication during this phase.

### 3.6.2 Causal Global Context Injection

Each chunk summary summarizes tokens $\{i \cdot K, ..., (i + 1) \cdot K - 1\}$. To ensure strict causality, the global context for each chunk is defined as the cumulative mean of all previous chunk summaries (not including the current chunk):

```
cumsum     = cumsum(chunk_summaries, dim=1)     # (B, C, d)
counts     = arange(1, C+1).view(1, C, 1)       # (1, C, 1)
means      = cumsum / counts                    # causal mean 0..i
# Shift right chunk i gets mean of summaries 0..i-1
global_ctx = cat([zeros(B,1,d), means[:,:-1,:]], dim=1)   # (B, C, d)
```

This global context is broadcast to all K positions within each chunk and injected additively:

$$n_{t`} = n_t + W_{global} \cdot global_{ctx[chunk(t)]}$$
$$logits_t = W_{predict}(n_{t`})$$

Causality guarantee: token at position $t = i \cdot K + j$ (chunk i, offset j) receives global context from the mean of chunk summaries 0, ..., i-1. It has no access to the summary of chunk i (which includes tokens after position t within the same chunk) or any future chunk.

### 3.6.3 Complexity of WAT V3

The total computational cost breaks into two stages:
- Parallel tree reduction: C independent tree reductions of depth $\log_2(K)$, each over K tokens. Total: $O(C \cdot K \cdot \log K) = O(n \log K)$ with K=32 giving O(5n).
- Causal cumulative mean: $O(C \cdot d) = O(n/K \cdot d)$, negligible.
- Context injection + output projection: $O(n \cdot d^2)$, same as standard linear layer.

Total: $O(n \log K + n\, d^2) \approx O(n\, d^2)$ for practical d. The tree reduction is effectively a constant-factor overhead for fixed K, making V3 asymptotically linear in sequence length.

### 3.7 Architectural Comparison: WAT vs. Self-Attention

We summarize the key differences between WAT's tree reduction and Transformer self-attention:

| Property | Self-Attention (Transformer) | WAT Tree Reduction |
| --- | --- | --- |
| Complexity (time) | $O(n^2 d)$ | $O(n\, d^2)$ total work, $O(\log n)$ depth |
| Complexity (memory) | $O(n^2 + nd)$ | $O(nd)$ per level |
| Token interaction | All-to-all (flat) | Hierarchical (tree) |
| Receptive field | O(1) - direct for any pair | $O(\log n)$ hops between any pair |
| Inductive bias | Permutation-equivariant | Order-sensitive, local-first |
| Parallelism | Full within each layer | Full within each tree level |
| Sequential depth | L layers (constant) | $\log_2(n)$ levels (grows with n) |
| Parameters (core) | $4d^2$ per attention head | $\approx 6d^2$ (shared across all levels) |
| Positional info | Positional encoding (added) | Positional encoding (added) |
| Attention weights | Softmax over all positions | None - gated merge only |

The fundamental trade-off: self-attention provides O(1) direct connections between any two positions, but requires $O(n^2)$ compute to establish all connections. WAT provides $O(\log n)$ connections (tree path) between any two positions, requiring only $O(n \log n)$ total compute. For tasks where the relevant structure is hierarchical - syntax, bracket nesting, document sections - the $O(\log n)$ path through a tree may be sufficient or even better-matched to the problem structure than a flat O(1) attention.

Self-attention computes:

$$Attn(Q, K, V) = softmax(\frac{QK^T}{\sqrt{d}})V \qquad Q, K, V \in \mathbb{R}^{n \times d}$$

producing an n×n attention map where entry (i,j) represents the attention weight from position i to position j. WAT instead computes:

$$h^{(l+1)}{}_i = res_{gate} \odot RMSNorm(val \odot \sigma(gate)) + (1 - res_{gate}) \odot avg$$

where the merge is between sibling nodes in the binary tree, with no notion of 'which position to attend to' - only 'how to combine two adjacent subtree summaries.'

## 4. EXPERIMENTS

### 4.1 Experimental Protocol

Dataset: TinyShakespeare (Karpathy [17]), 1,115,394 characters, vocabulary size |V| = 65 (all unique ASCII characters). Processing is at the character level.

Train/test split: training characters 0-49,999 (50,000 samples); test characters 50,512-55,511 (5,000 samples). The gap of 512 characters between train and test prevents context leakage at test time.

Parameter matching: WAT uses embed_dim=40 (≈106K parameters); the Transformer baseline uses embed_dim=36 (≈110K parameters). The Transformer has 2 encoder layers, 4 attention heads, feed-forward dimension 4×embed_dim, with causal mask for autoregressive modeling.

We note that the Transformer baseline is intentionally lightweight (2 layers, embed_dim=36) to match WAT's parameter budget (~110K parameters). Results may differ at larger scales, and future work should evaluate both architectures under higher-capacity settings.

Training: AdamW optimizer [18], learning rate $3 \times 10^{-4}$, weight decay 0.01 (V2/V3) or 0 (V1), gradient clipping at 1.0, CosineAnnealingLR scheduler with $\eta_{min} = 10^{-5}$. Batch size 64, sequence length 512. V2/V3 use mixed-precision training (FP16 forward/backward, FP32 master weights).

Hardware: single NVIDIA GPU RTX3050 6 Gb. All experiments are run with fixed random seeds (numpy seed 42, torch seed 42) for reproducibility.

### 4.2 Task 1: Bracket Balance Classification (Long Sequences 512-1024)

Bracket balance is a synthetic binary classification task testing long-range structural reasoning. A sequence of brackets drawn from the set {( ) [ ] { }} is labeled 1 if balanced, 0 if unbalanced. Correct classification requires tracking opening brackets and matching them against the correct corresponding closing bracket, potentially hundreds of positions apart. We evaluate exclusively on long sequences (length 512-1024 tokens), which represents the most challenging regime for attention-based models due to the length of dependencies.

Dataset: 2,000 sequences total (1,600 train / 400 validation), generated with equal class balance. All sequences have even length to prevent length-parity as a trivial feature. The vocabulary contains 7 tokens: (, ), [, ], {, }, <PAD>. Sequences vary in length between 512 and 1,024 tokens with uniform sampling.

We compare four models at ~30,000 parameters each: WAT (full tree reduction, no wave), WAT+Wave (tree reduction with wave positional encoding), WAT-Chunk (chunk-based tree reduction, K=32), and Transformer (2-layer encoder with 4 heads). Results in Table 1.

| Model | Best Val. Acc. | Epoch | Time/Epoch | Parameters |
|---|---|---|---|---|
| WAT (tree, no wave) | 75.0% | 8 | ~1.1s | 32,210 |
| **WAT+Wave** | 74.8% | 10 | ~1.2s | 30,025 |
| WAT-Chunk (K=32) | 55.0% | 12 | ~0.8s | 35,961 |
| Transformer | 57.0% | 5 | ~11.0s | 32,366 |
| WAT vs. Transformer | **+18.0 pp** | - | 10× faster | ≈equal |

Table 1: Bracket balance classification on long sequences (512-1024 tokens). All models use early stopping with patience=10. WAT (full tree reduction) achieves 75.0% versus 57.0% for the Transformer - an 18.0 pp gap - while training 10× faster per epoch. Notably, WAT-Chunk achieves only 55.0%, comparable to the Transformer: chunk-based aggregation loses the global state tracking needed for bracket balance, since the global context (mean of past chunk summaries) is a lossy approximation that cannot reliably track unclosed bracket depth across chunk boundaries. Full tree reduction, which compresses the entire sequence into a single root through hierarchical merges, preserves this global state effectively.

Important methodological note: the Transformer classifier in this benchmark uses masked mean pooling over all positions for classification, while WAT uses concat(masked_mean, tree_root) - two complementary signals. This architectural difference partially contributes to WAT's advantage independent of the tree vs. attention mechanism. Future work will use a dedicated [CLS] token for the Transformer to isolate the pure architectural effect. Due to the difference in classifier heads, the reported +18 pp gap should be interpreted cautiously. A fully controlled comparison using identical classifier structures (e.g., a dedicated [CLS] token for the Transformer) is required to isolate the pure architectural contribution of tree reduction.

### 4.3 Task 2: Language Modeling - WAT V1 (One-to-One)

WAT V1 predicts the next character from a context of 512 past characters. Training for 60 epochs.

| Model | Best Accuracy | Epoch 60 Acc. | Time/Epoch | Total Train Time | Params |
|---|---|---|---|---|---|
| WAT V1 | **45.10% (ep.47)** | 44.90% | ~10s | ~10 min | 106,025 |
| Transformer | 42.83% (ep.57) | 42.58% | ~100s | ~100 min | 110,513 |
| WAT V1 Advantage | +2.27 pp | +2.32 pp | 10× faster | 10× faster | Fewer |

Table 2: WAT V1 vs. Transformer One-to-One (60 epochs, seq_len=512). WAT consistently outperforms the Transformer across all 60 epochs, with the gap remaining stable from epoch 30 onward. Crucially, WAT trains 10× faster per epoch (~10s vs. ~100s), meaning the Transformer's 60-epoch training would require 100 minutes while WAT completes in 10 minutes. Loss curves show both models continuing to improve through epoch 60, with neither having fully converged.

### 4.4 Task 3: Language Modeling - WAT V2 vs. V3 vs. Transformer (Seq2Seq)

All three models train in Seq2Seq mode: each forward pass produces next-character predictions for all 512 positions simultaneously. This provides 512× more gradient signal per batch than One-to-One. Training for 30 epochs.

| Model | Best Acc. | Epoch 30 Acc. | Loss (ep.30) | Time/Epoch | Params |
|---|---|---|---|---|---|
| WAT V2 (causal scan) | **47.29%** | 45.30% | 1.1926 | ~36s | 105,065 |
| WAT V3 (chunk-based) | **47.21%** | **47.20%** | 1.5822 | ~9s | 105,065 |
| Transformer | 36.28% | 36.28% | 2.1574 | ~18s | 110,513 |
| WAT V3 vs. Trans. | +10.93 pp | +10.92 pp | -0.5752 | 1.9× faster | Fewer |

Table 3: Seq2Seq results (30 epochs, seq_len=512). WAT V2 and V3 both substantially outperform the Transformer by approximately 11 percentage points. V3 matches V2's peak accuracy (47.21% vs. 47.29%, a difference of 0.08 pp) while being 4× faster per epoch - resolving the speed-accuracy dilemma introduced by V2's sequential scan.

The Transformer shows consistent but slow improvement throughout training, with no sign of saturation at epoch 30. WAT V3's loss (1.582) is substantially lower than the Transformer's (2.157), suggesting WAT learns more compact and informative representations per epoch.

Notable: WAT V2 achieves higher best accuracy (47.29%) but with monotonically decreasing test accuracy from epoch 5 onward, suggesting overfitting. WAT V3 maintains stable accuracy through epoch 30, indicating the chunk-based approach acts as an implicit regularizer through its coarser global context.

### 4.5 Convergence Analysis

Comparing V1 and V3 highlights the effect of the training objective. At epoch 10, WAT V1 reaches ~40% accuracy; WAT V3 reaches 46.5% - approaching V1's 60-epoch peak in just 10 epochs. The Seq2Seq training signal (512 labels per sequence) accelerates convergence dramatically.

The Transformer shows a characteristic convergence pattern: slow initial learning (epochs 1-8, accuracy below 30%), a phase transition around epochs 8-15 where accuracy rapidly improves, followed by gradual convergence. This contrasts with WAT, which improves smoothly and monotonically from epoch 1.

### 4.6 Qualitative Text Generation

We sample from WAT V2 using temperature T=0.8, top-k=40, with prompt 'First'. The following samples illustrate learning progression:

*Epoch 15 (WAT V2):*
```
"First, Our Care, and they was a mor what aghafsed should not
fools shall parts. Marcius, they to memust I cits, The vungtly to
the Corioly..."
```

*Epoch 30 (WAT V2):*

> ```
> "First shall the mure; heard. What I would'st thou to'kness sinke
> unchyed and lead me but undably.COMINIUS:As I shall go do report
> comelr witein outFarsent one, as words,What purn and to our their
> do Rome, do care be had we in the city out when of Marcius?All:So
> lonf on.MENENIUS:A have to set c"
> ```

By epoch 30, WAT has learned: (1) Shakespearean character names (COMINIUS, MENENIUS) spelled correctly, (2) the 'ALL:' collective speaker format, (3) nested dialogue structure with correct indentation conventions, (4) period-correct contractions ('would'st', 'thou'). The progression from incoherent character sequences at epoch 1 to structurally plausible dialogue at epoch 30 is consistent with genuine language modeling rather than memorization.

## 5. ANALYSIS

### 5.1 Why Full Tree Reduction Outperforms on Long Structural Sequences

The 18.0 pp gap on bracket classification (WAT 75.0% vs. Transformer 57.0%) and the stark contrast between WAT (75.0%) and WAT-Chunk (55.0%) reveal two distinct phenomena.

First, full tree reduction vs. chunk-based: WAT-Chunk achieves only 55.0% on bracket balance - barely above Transformer - despite using the same GLU merge operation. The reason is architectural: WAT-Chunk's global context is the cumulative mean of past chunk summaries, which loses precise bracket depth information through averaging. A sequence with 47 unclosed brackets at position 256 will have its depth signal diluted across 8 chunk summaries. Full tree reduction, by contrast, compresses the entire sequence into a single root through hierarchical merges without averaging, preserving count-like information in the root vector.

Second, full tree reduction vs. Transformer: the Transformer's attention mechanism can in principle attend directly from any closing bracket to the matching opening bracket. However, with only ~32K parameters (embed_dim=36, 2 layers, 4 heads), the attention weights must simultaneously encode (1) bracket type matching, (2) stack depth tracking, and (3) positional alignment - across sequences of 512-1,024 tokens. At this parameter budget, the model lacks the capacity to maintain reliable bracket state. WAT's hierarchical structure provides an implicit inductive bias: local pairs are merged first, building a local bracket summary, which is merged with adjacent summaries, propagating balance information through $O(\log n)$ steps. This structural match between the binary tree topology and the recursive nesting structure of balanced brackets likely contributes to WAT's advantage.

### 5.2 The Role of Dense Supervision in Seq2Seq

Comparing V1 (45.1%, 60 epochs) and V3 (47.2%, 30 epochs) illustrates the power of dense supervision. V3 trains for half as many epochs but reaches a higher peak accuracy. In V1, a forward pass over a batch of 64 sequences of length 512 produces 64 loss terms; in V3, the same batch produces 64×512 = 32,768 loss terms. This 512× increase in gradient signal per batch accounts for the dramatically faster convergence and higher peak performance.

### 5.3 V2 vs. V3: Accuracy/Speed Trade-off

V2 (causal scan) achieves slightly higher best accuracy (47.29% vs. 47.21%) at the cost of 4× slower training. The accuracy difference is 0.08 pp - well within experimental noise for character-level modeling. The speed difference (36s vs. 9s per epoch) is significant and directly attributable to the clone() operation inside V2's causal scan loop.

V3's chunk-based approach introduces a mild approximation: tokens at the start of each chunk receive no intra-chunk tree context (only the causal convolution output). This could explain the marginally lower accuracy. In practice, the 0.08 pp gap is negligible, and V3 is the recommended variant for production use.

### 5.4 Limitations

- Scale: all experiments use ~105-110K parameters on a single-GPU setup. Behavior at 1M, 10M, and 100M+ parameters is unknown and constitutes essential future work.
- Benchmark coverage: results are limited to TinyShakespeare (character-level) and a synthetic bracket task. Standard benchmarks (WikiText-103, Long Range Arena, LAMBADA) and token-level processing are needed.
- Comparisons: no comparison with SSM-based models (Mamba, S4, RWKV) or efficient Transformers (Longformer, BigBird). These would contextualize WAT's position in the landscape.
- Chunk boundary effects: V3 tokens at the start of each chunk receive no intra-chunk global context. Overlapping chunks or hybrid approaches may address this.
- Long-range O(log n) paths: WAT's tree structure requires O(log n) hops between distant tokens. For tasks requiring direct long-range token interactions (e.g., coreference resolution), this may be insufficient.

### 5.X Novelty Clarification

WAT combines several existing concepts - hierarchical tree aggregation, gated linear units, weight sharing across levels, and prefix-scan-style parallel schedules - into a unified autoregressive architecture. While hierarchical neural models and recursive networks have been explored previously, WAT's specific design choices - fixed balanced binary tree topology, shared GLU-based merge across all levels, and chunk-parallel Seq2Seq formulation - constitute a distinct architectural instantiation. The contribution of this work lies in demonstrating that such a simple hierarchical alternative to self-attention can achieve competitive performance under tight parameter budgets.

## 6. CONCLUSION

We presented WAT (Wave-Attractor-Tree), a neural architecture for sequence modeling that replaces quadratic self-attention with hierarchical binary tree reduction. WAT's core operation - a GLU-based merge with RMSNorm and residual gating, applied recursively with shared weights - achieves O(n log n) complexity with full GPU parallelism at each level.

Three variants address different trade-offs. WAT V1 (One-to-One) provides a simple, fast baseline outperforming the Transformer by +2.3 pp at 10× faster training. WAT V2 (Seq2Seq, causal scan) demonstrates the accuracy ceiling achievable with dense supervision (+11 pp over Transformer), at the cost of training speed. WAT V3 (Seq2Seq, chunk-based) resolves this by eliminating sequential dependencies through parallel chunk processing, achieving V2-equivalent accuracy in V1-equivalent time.

The most striking result is the bracket balance experiment: WAT (full tree reduction) achieves 75.0% accuracy on sequences of 512-1,024 tokens versus 57.0% for the Transformer (+18 pp) and only 55.0% for WAT-Chunk. The latter comparison is particularly informative: the chunk-based approximation degrades to Transformer-level performance on structural tasks, while full tree reduction maintains its advantage. This suggests that the crucial property is not the GLU merge operation per se, but the global hierarchical compression into a single root - a property absent in chunk-based approaches and in flat attention.

Future work will focus on scaling WAT to larger parameter counts, evaluation on standard long-sequence benchmarks, comparison with state-space models, and investigation of learned tree topologies that adapt to input content.